\documentclass[oribibl]{llncs}
\usepackage{multicol}
\usepackage{amsmath}
\usepackage{amssymb}
\usepackage{graphicx}
\usepackage{epsfig}
\usepackage{color}
\usepackage{hhline}

\newcommand{\ignore}[1]{}

\ignore{+++
\newcounter{theorem-counter}
\newcounter{corollary-counter}
\newcounter{lemma-counter}
\newcounter{definition-counter}
\newcounter{example-counter}
\newcounter{proposition-counter}
\newcounter{remark-counter}

\setcounter{theorem-counter}{0}
\setcounter{corollary-counter}{0}
\setcounter{lemma-counter}{0}
\setcounter{definition-counter}{0}
\setcounter{example-counter}{0}
\setcounter{proposition-counter}{0}
\setcounter{remark-counter}{0}

{\vskip \abovedisplayskip \refstepcounter{theorem-counter}%
\noindent {\bf Theorem \arabic{theorem-counter}.}}%
{\boxtheorem}

{\vskip \abovedisplayskip \refstepcounter{corollary-counter}%
\noindent {\bf Corollary \arabic{corollary-counter}.}}%
{\boxtheorem}

{\vskip \abovedisplayskip \refstepcounter{lemma-counter}%
\noindent {\bf Lemma \arabic{lemma-counter}.}}%

\newenvironment{definition}%
{\vskip \abovedisplayskip \refstepcounter{definition-counter}%
\noindent {\bf Definition \arabic{definition-counter}.}}%
{\newline}

\newenvironment{example}%
{\vskip \abovedisplayskip \refstepcounter{example-counter}%
\noindent {\bf Example \arabic{example-counter}.}}%

\newenvironment{proposition}%
{\vskip \abovedisplayskip \refstepcounter{proposition-counter}%
\noindent {\bf Proposition \arabic{proposition-counter}.}}%

{\vskip \abovedisplayskip \refstepcounter{remark-counter}%
\noindent {\bf Remark \arabic{remark-counter}.}}%

+++}

\ignore{++
\usepackage{amsmath}
\usepackage{amsthm}
\usepackage{amssymb}
\usepackage{mathtools}
\usepackage{nicefrac}
\usepackage[linesnumbered,algoruled,commentsnumbered]{algorithm2e}
\usepackage{todonotes}
\allowdisplaybreaks
\usetikzlibrary{positioning}
++}

\newcommand{\ul}[1]{\underline{#1}}
\newcommand{\boxtheorem}{\hfill $\blacksquare$\\}
\newcommand{\nit}[1]{{\it #1}}

\ignore{+++
\newcounter{propositionA-counter}
\newcounter{lemmaA-counter}

\setcounter{lemmaA-counter}{0}
\setcounter{propositionA-counter}{0}

{\vskip \abovedisplayskip \refstepcounter{lemmaA-counter}%
\noindent {\bf Lemma A.\arabic{lemmaA-counter}.}}%

{\vskip \abovedisplayskip \refstepcounter{propositionA-counter}%
\noindent {\bf Proposition A.\arabic{propositionA-counter}.}}%

++}

\newcommand{\mc}[1]{\mathcal{ #1}}

\newcommand{\mbf}[1]{\mathbf{ #1}}

\newcommand{\mf}[1]{\mathfrak{ #1}}

\newcommand{\e}{\mathbf{e}}
\newcommand{\C}[1]{\mathcal{C}}
\newcommand{\T}[1]{\mathcal{T}}

\newcommand{\red}[1]{\textcolor{red}{#1}}
\newcommand{\re}[1]{\textcolor{red}{#1}}

\newcommand{\bl}[1]{\textcolor{blue}{#1}}

\ignore{+++++

\renewenvironment{thebibliography}[1]{
  \begin{oldthebibliography}{#1}
    \setlength{\itemsep}{0em}
    \setlength{\parskip}{0em}
}
{
  \end{oldthebibliography}
}

\setcounter{secnumdepth}{3}

+++}

\title{{\bf Reconciling Consistency-Based Diagnosis with Actual-Causality-Based Explanations}\ignore{ for Each Other's Benefit}\thanks{{\bf Dedicated to the memory of Joseph Y. Halpern, a great scholar, a \linebreak universal researcher.}}}

 \author{\bf Leopoldo Bertossi\thanks{Emeritus Professor. Adjunct Professor Western University, Canada. bertossi@scs.carleton.ca}\\Carleton University, Canada \  \& \ IMFD, Chile}

\date{}

\institute{}

\begin{document} \pagestyle{plain} \maketitle \thispagestyle{empty}

\begin{abstract} We establish, from the point of view of Explainable AI (XAI), connections between Consistency-Based Diagnosis (CBD), on one side, and Actual Causality and Causal Responsibility, on the other. CBD has received little attention from the XAI community. Connections between these two areas could have a fruitful impact on   XAI and Explainable Data Management.
\end{abstract}

\section{Introduction}\label{intro2}

 When we have a system $\mc{S}$ that receives an input $I$, and returns an output $O$, it may be the case that, for that particular input, $\mc{S}$ returns the wrong output. We are then confronted with {\em uncertainty} about what is wrong with $\mc{S}$, or, more precisely, about what of $\mc{S}$'s {\em components} is not working properly. If we have a model of the system, we may try to identify them. This is the subject of {\em model-based diagnosis}.

  On the other side, if $\mc{S}$ returns an output $O^\star$ for an input $I^\star$, we are commonly {\em uncertain} about what caused output $O^\star$. Assuming that we  know what the system returns in general, and also that the system is working properly,  we may wonder what component of $I^\star$ determined $O^\star$ the most. This is the subject of causality; {\em actual causality} in our case.

Explainable AI (XAI), and in particular Explainable Machine Learning, have become relevant areas of research in AI. {\em Actual Causality}, first proposed by Joe Halpern and Judea Pearl \cite{HP05}, and {\em Causal Responsibility}, first proposed by Hana Chockler and Joe Halpern \cite{halpernChockler}, have been applied in XAI, to provide explanations for outcomes from machine-learning (ML) models \cite{deem,TPLP22,izza}. They have also been applied in Explainable Data Management (XDM), to provide explanations for query answers \cite{suciu,tocs,flairsExt}.  See
\cite{bda22} for a survey of some approaches. \ 
Causality has gained prominence in  ML, both for explaining and interpreting learned models, but also for learning itself \cite{janzing,schoelkopf,schoel}.

  A more classic area  of AI is {\em Model-Based Diagnosis} (MBD) \cite{struss}. It deals with  providing diagnosis and explanations for a system's behavior on the basis of a model of the system, usually of the  kind of models found the area of  Knowledge Representation in AI. 
  
  One of the  prominent forms of MBD, {\em Consistency-Based Diagnosis} (CBD), proposed by Ray Reiter \cite{reiterDiag}, is typically applied to obtain {\em diagnoses} for a system that exhibits an unintended behavior.  CBD has an interesting role to play in XAI. Abduction, or sufficient explanations, is another form of MBD that has found its way into XAI \cite{marquis,darwiche,joao,izza}; and also in XDM \cite{flairsExt,bienvenu,nina}. Since there is larger body of recent research on the use of abduction in XAI , we do not consider abduction in this work. CBD has received much less attention.

 In this work we explore  connections between actual causality and CBD. They  haven't received much attention; and we think the two areas can profit from each other. Actually, in \cite{tocs}, an early connection allowed us to design algorithms for obtaining causal explanations in XDM, and to obtain complexity results for CBD. For reasons of space, and to best convey intuitions, we unveil and formulate interesting connections  by means of examples. However, it should become clear how to formulate things in general terms. We also show how the established connections can be exploited in technical terms. Exploring them in more depth is is part of ongoing work.

  In this work we stick to the propositional case, that is, logical specifications, features, classifiers, and models on which CBD is performed are all written in propositional logic or are Boolean.

 In Section \ref{sec:causality}, we review and present actual causality and responsibility as used to explain outcomes from classification models, those usually learnt in ML. In Section \ref{sec:cbd}, we show how a typical problem of CBD can be recast as one of actual causality, and how responsibility could become a new ingredient in CBD. On the basis of CBD problems, we give a precise definition of actual causality, and responsibility. Furthermore, we show how the CBD problem could be represented by means on a Structural Causal Model \cite{pearl}. In Section \ref{sec:diagX}, we proceed the other way around. We take a classification problem that would be normally approached via actual causality and responsibility (or some other attribution method of XAI), and formulate it as a CBD problem. In Section \ref{sec:connections}, we show an example of how CBD can borrow techniques from  actual causality, and the other way around.

\section{Actual Causality and Responsibility in XAI}\label{sec:causality}

The goal of this section is twofold. We describe and illustrate, by means of an example, the main concepts of {\em actual causality}; and we also show how to apply them in XAI, in this case, to obtain explanations from a ML-based classifier. Precise definitions in the context of CBD are given in Section \ref{sec:cbd}.

The basic idea behind  {\em actual causality} \cite{HP05,H16} is that of performing {\em counterfactual interventions} on the values of variables of a model, changing their values in order to detect if there are changes in other variables, typically the output of the model. The interventions lead to {\em counterfactual and actual explanations}.

\begin{example}\label{ex:ml}
Consider a classifier $\mc{C}$ an in Figure \ref{fig:class0}(a) that has been learned from training data. After that, it can be too complex; and we may not  have an idea of what is going on inside (think of a huge neural network). So, $\mc{C}$ may well be (or treated as) a black-box classifier, but  still we can use $\mc{C}$'s input/output relation. For an input entity $\e = \langle \mbf{x}_1,\mbf{x}_2,\mbf{x}_3,\mbf{x}_4\rangle \in \{0,1\}^4$, with four binary feature values, the label $L(\e) \in \{0,1\}$ is returned.\vspace{-3mm}

\begin{figure}[h]
\centerline{\includegraphics[width=5cm]{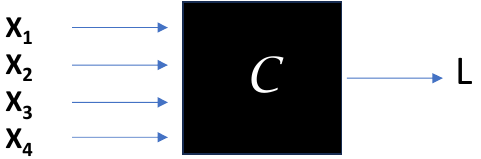}~~~~~~~~~~\includegraphics[width=5.4cm]{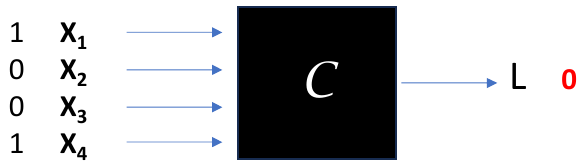}}
\vspace{-2mm}\caption{(a) A Binary Classifier. \ \ \ \ \ \ (b) Particular Input/Output.}\label{fig:class0}
\end{figure}\vspace{-3mm}

As shown in Figure \ref{fig:class0}(b), with input $\e = \langle 1,0,0,1\rangle$, we obtain label $0$. We want an explanation, in the form of feature values in $\e$ that are actual causes for the outcome.
Do changes of (or interventions on) feature values of $\e$ change the label from $0$ to $1$?

\begin{figure}[h]
\centerline{\includegraphics[width=5.4cm]{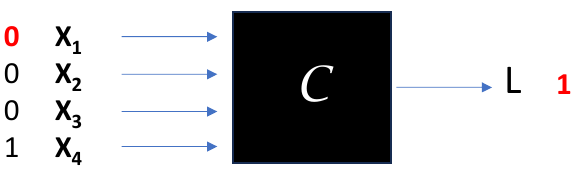}~~~~~~~~~~\includegraphics[width=5.4cm]{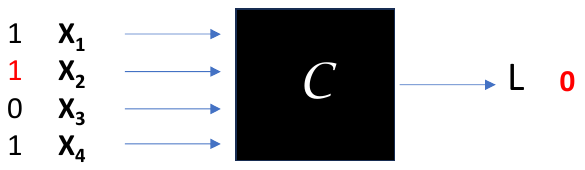}}
\vspace{-2mm}\caption{(a) Successful Intervention. \ \ \ \ \ \ (b) Unsuccessful Intervention.}\label{fig:class1}
\end{figure}\vspace{-3mm}

As shown in Figure \ref{fig:class1}(a), the change on $\mbf{x_1}$ from $1$ to $0$ changes the label. The value $\mbf{x_1} =1$ is said to be a {\em counterfactual cause} for the initial label:  A change on $\mbf{x}_1$ alone already changes the label.

Now, let us concentrate on  $\mbf{x_2}$. Figure \ref{fig:class1}(b) shows that $\mbf{x}_2=0$ is not a counterfactual cause for $L=0$. Let's keep its value for a moment, and consider the following unsuccessful intervention on $\mbf{x}_3, \mbf{x}_4$ in Figure \ref{fig:class2}(a).\vspace{-3mm}

\begin{figure}[h]
\centerline{~~\includegraphics[width=5.4cm]{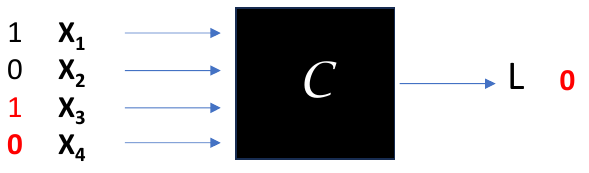}~~\includegraphics[width=7cm]{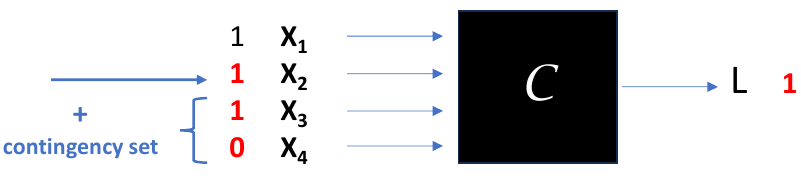}}
\vspace{-3mm}\caption{(a) Unsuccessful Intervention.  \ \ \ \ \ \ \ \  (b) Successful Intervention.}\label{fig:class2}
\end{figure}

\vspace{-2mm}If, in addition to those two changes, we change again $\mbf{x}_2$ to $1$, we are now successful, as shown in Figure \ref{fig:class2}(b). We say that $\mbf{x}_2 =0$ is an {\em actual cause} for the original label, and $\Gamma = \{\mbf{x}_3=0, \mbf{x}_4 = 1\}$ is a {\em contingency set} (CS) for $\mbf{x}_2 =0$: In order for the change on $\mbf{x}_2$ to change the label, it needs an additional, contingent set of changes. However, those two contingent changes alone do not change the label. \ We say that $\e^\prime = \langle \re{0},0,0,1\rangle$ and $\e^{\prime\prime} = \langle 1,\re{1},\re{1},\re{0}\rangle$ are {\em counterfactuals} (counterfactual versions) of the original entity $\e$.\boxtheorem
\end{example}

\vspace{-3mm}The {\em causal responsibility}  of an actual cause  is a numerical quantification of its strength as a cause. It is based on the number of additional changes an actual cause needs to change the label. Actually, for an actual cause $\mbf{x} = v$, its responsibility is defined by $\nit{Resp}(\mbf{x}) := 1/(1+ |\Gamma|)$, where $\Gamma$ is a minimum-size CS for $\mbf{x}$. The responsibility of a feature value that is not an actual cause is $0$, by definition.

\begin{example}\label{ex:ml2} (ex. \ref{ex:ml} cont.) \ Since $\mbf{x}_1 =1$ in $\e$ does not need additional changes, $\Gamma = \emptyset$ is its minimum CS, and then, its responsibility is $\nit{Resp}(\mbf{x}_1) \ = \ 1$, the maximum possible responsibility.
\ If we assume that $\Gamma = \{\mbf{x}_3=0, \mbf{x}_4 = 1\}$, as above, is a minimum CS  for $\mbf{x}_2 = 0$,  $\nit{Resp}(\mbf{x}_2)  = 1/(1 + 2) = 1/3$.
 \boxtheorem \end{example}

\vspace{-2mm}
Actual causality provides counterfactual explanations to observations. In general
terms, they are “components” of a system that are a cause for an observed behavior.
Counterfactual causes are actual causes with an empty CS. Accordingly, counterfactual causes are {\em strong} causes in that they, by themselves, explain the observation. Actual causes that are not counterfactual causes are {\em weaker} causes, they require the company of other components to explain the observation.

In some applications of actual causality, there may be variables that are declared {\em endogenous}, while the others are {\em exogenous}. The former are of interest for causality purposes. In particular, only endogenous variables can be actual causes and members of CSs. In our  example, we could have declared $\mbf{x}_3$ as exogenous; for example, if we know that its value has to be $0$, no matter what. This value is not subject to interventions. In that case, $\mbf{x}_3$ could not have been a member of the CS any longer. We would have to look for contingencies somewhere else. The choice of endogenous/exogenous variables is application dependent. See \cite[sec. 5]{pearl} for a relevant discussion.

Actual causality and responsibility can be applied without necessarily knowing ``the internals" of the classifier, which can be (or be treated as) a ``black box". Only the input/output relation is needed.
\ Responsibility has become, as more generally called in XAI, an {\em attribution score}. The kind of explanations obtained are {\em local} in that they apply to values in a single input entity. As defined, responsibility does not provide a {\em global} explanation for the behavior of the classifier.

The computation of the responsibility score in explainable ML is known to be  intractable, already in the binary case \cite{TPLP22}. Above we introduced responsibility in a binary setting. When feature values are not binary, its definition has to be extended. This was done, analyzed and experimented with in \cite{deem} (see also \cite{adbis23}).

\section{Consistency-Based Diagnosis as Actual Causality}\label{sec:cbd}

We will first introduce and illustrate {\em consistency-based diagnosis} (CBD)  \cite{reiterDiag,ray92} by means of an example. After that, we will show how to cast it as an actual causality problem related to classification.

 If we are confronting a system that is exhibiting an unexpected or abnormal behavior, we want to obtain a {\em diagnosis} for this, i.e. some sort of explanation. Diagnoses are obtained from a model of the system.

 \begin{example} \label{ex:circ} Figure \ref{fig:circ}(a) shows a very simple Boolean circuit with an {\em And}-gate, $A$, and an {\em Or}-gate, $O$. The input variables are $a,b,c$, the intermediate output variable for $A$ is $x$, and the final, output variable is $d$; all of them taking values $0$ or $1$. The intended meaning of the propositional variable $a$ is ``input $a$ is true" (or takes value $1$), etc. With the indicated inputs and output, the circuit seems to be behaving as expected.\vspace{-3mm}

\begin{figure}[h]
 \centerline{\includegraphics[width=5.5cm]{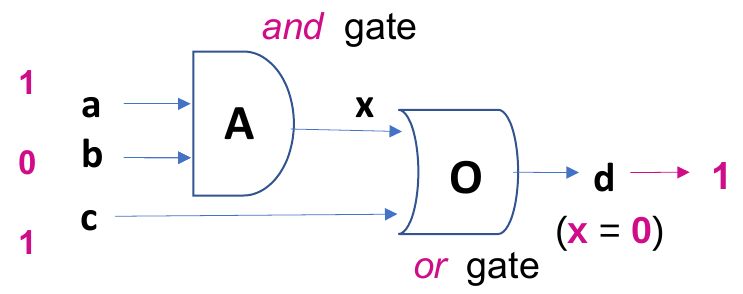}~~~~~~\includegraphics[width=6cm]{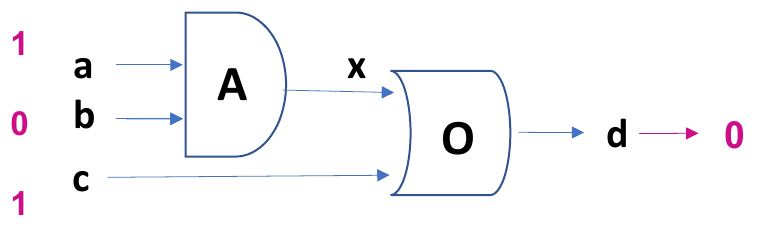}}\vspace{-1mm}
 \caption{(a) Boolean Circuit. \ \ \ \ \ \ \ \ \ \ (b) Faulty Boolean Circuit.}\label{fig:circ}\vspace{-3mm}
\end{figure}

Figure \ref{fig:circ}(b) shows an unintended behavior: with those inputs, the output should be $1$.
We need a {\em diagnosis} for the abnormal behavior of the circuit.
\ The define a diagnosis, we need a model of the system.

In this example,
a logical model of the circuit, when it works properly, is a set of propositional formulas:
$\{(x \longleftrightarrow (a \wedge
b)),  \ (d \longleftrightarrow (x \vee
c))\}$.  However, our circuit at hand, by working abnormally, is {\em not} modeled by these formulas. \
Furthermore,
the {\em observation}, $\nit{Obs} = \{ a, \neg b, c, \neg d\}$, indicating that $a$ and $c$ are true, but $b$ and $d$ are  false, is mutually inconsistent with this ideal model: there is no assignment of truth values to the propositional variables that makes the combination true. From their combination we cannot logically obtain any useful information.

We may want instead {\em a model that allows failures, or abnormal behaviors}. From such a model, we could try to obtain explanations for them.
\ A better, more flexible model  that allows failures, and specifies how components behave under normal conditions is:
\begin{equation}\mc{M} = \{\neg abA \longrightarrow (x \leftrightarrow (a \wedge
b)), \ \ \neg abO \longrightarrow (d \leftrightarrow (x \vee
c))\}. \label{eq:M}
\end{equation}

The first formula says
{\em ``When $A$ is not abnormal,  it works as an
And-gate"}, etc. Here, $\nit{abA}$ and $\nit{abO}$ are new propositional variables.
This is a ``weak model of failure" in that it specifies how things behave under normal conditions, but not under abnormal ones. This is a common kind of models under CBD \cite{ray92}.

Model $\mc{M}$ assumes that the only potentially faulty components are, in this case, the gates (but not the wires connecting them), a modeling choice. \
\ Now gates could be abnormal (or faulty), and $\nit{Obs} \ \cup \ \mc{M}$ \ is a perfectly consistent set of formulas.
When one specifies, in addition, that the gates are not abnormal, we reobtain that:
\begin{equation}
  \{a, \neg b, c, \neg d\} \ \cup \ \mc{M} \ \cup \{\neg \nit{abA}, \neg \nit{abO}\} \ \mbox{ is \ inconsistent}. \label{eq:inco}
  \end{equation}

\noindent Then, something has to be abnormal, and it is through {\em consistency restoration} that we will be able to characterize and compute diagnoses. \ If we make gate $\bl{O}$ abnormal in (\ref{eq:inco}), we restore consistency, that is, in contrast to (\ref{eq:inco}): \ 
$\nit{Obs} \ \cup \ \mc{M} \ \cup \{\neg \nit{abA}, \ \ul{\nit{abO}}\} \ \mbox{ is \ consistent}$ (underlining the change  made). Accordingly,  and by definition, $\Delta = \{\nit{abO}\}$ is a diagnosis. \ Similarly,
 $\Delta^\prime = \{\nit{abO}, \nit{abA}\}$ is a diagnosis, because making every gate abnormal also restores consistency:
$
\nit{Obs} \ \cup \ \mc{M} \ \cup \{\ul{\nit{abA}}, \ \ul{\nit{abO}}\} \ \mbox{ is \ consistent}$.

 We may consider $\Delta$ as a ``better" diagnosis than $\Delta^\prime$, because  it makes fewer assumptions; it is more informative by providing a narrower and more focused diagnosis. $\Delta$ is a {\em minimal diagnosis} in that it is not set-theoretically included in any other diagnosis. It is also a {\em minimum diagnosis} in that it has a minimum cardinality. \boxtheorem
\end{example}\vspace{-6mm}

\subsection{CBD as a Causal Problem}

Now, we are going to cast the CBD problem as an actual causality problem, as applied to a classification problem. We do this by appealing to our example.

\begin{example} \label{ex:circAC} (ex. \ref{ex:circ} cont.) \  Our classification problem is that of deciding if an input -with a model- is satisfiable or not. The model is $\mc{M}$ in (\ref{eq:M}). Our classifier, with its binary inputs and output, is shown in Figure \ref{fig:sat}. It consists of model $\mc{M}$ plus a SAT solver that verifies if $\mc{M}$ is consistent with the input  $\e$, whose last component, $d$, stands for the output of the circuit.

\vspace{3mm}
\begin{figure}[h]
\hspace*{7mm} $\e =  \langle abA, abO, a, b, c,d\rangle$

\hspace*{4.8cm}$\longrightarrow$

\hspace*{6mm} {\footnotesize (propositional~feature~vector)}

\vspace{-2.6cm}
\hspace*{0.5cm}\begin{align*}
SAT?~~~~~~~~~~~~~&\\
\hspace*{2.8cm}\boxed{
\begin{aligned}
&\neg abA \longrightarrow (x \leftrightarrow (a \wedge
b))\\
&\neg abO \longrightarrow (d \leftrightarrow (x \vee
c))\\
&~~~~~~~+ \ \mbox{ SAT solver}
\end{aligned}
}
\end{align*}

\vspace{-1.3cm}\hspace*{9.5cm} $\longrightarrow \ \nit{yes}/\nit{no}$
\vspace{6mm}\caption{A Consistency Classifier}\label{fig:sat}
\end{figure}

\vspace{-4mm}As we saw in Example \ref{ex:circ}, with $\e = \langle 0,0,1,0,1,0\rangle$, representing $\langle \neg \nit{abA},$ $ \neg \nit{abO}, a,$ $, \neg b, c, \neg d\rangle$, the classifier returns {\em no} (or $0$).
Then, we can apply actual causality by first declaring ``features" $\nit{abA}, \nit{abO}$ as endogenous, subject to interventions, while the others, $a,b,c,d$, are exogenous. The latter provide context, and are not subject to interventions.
\ By counterfactually intervening $\e$, producing $\e^\prime = \langle 0,\red{1},1,0,1,0\rangle$, indicating that $\nit{abO}$ is true, we manage to change the classifier's output to $\nit{yes}$. This does not happen with $\nit{abA}$ though. Then, ``$\nit{abO} =0$" is a counterfactual cause.   Intervening both $\nit{abA}$ and $\nit{abO}$ also changes the output, but this set of changes includes one that is already a counterfactual cause. Accordingly, $\nit{abA}$ is neither a counterfactual nor an actual cause.

Now we can apply responsibility to our CBD scenario: $\nit{Resp}(\nit{abO}) := 1$, and $\nit{Resp}(\nit{abA}) := 0$.\boxtheorem
\end{example}

\vspace{-2mm}
Our example suggests a correspondence between actual causes with their minimal/minimum contingency sets and minimal/minimum diagnoses. To formulate this connection, we proceed now to give a formal definition of actual cause as associated to a MBD problem. (In Section \ref{sec:causality}, we just provided an example, to convey the gist.)

\subsection{Revisiting the Connection}

Let us  start with a {\em weak model of failure} of a system, as in Example \ref{ex:circ}, that is, a specification in propositional logic, $\mc{M}$, of the form: \begin{equation}\label{eq:mod} \mc{M} = \{
\neg abC_1 \rightarrow \varphi_1, \
\neg abC_2 \rightarrow \varphi_2, \
\ldots
\neg abC_n \rightarrow \varphi_n\},\end{equation}

\vspace{-2mm}\noindent for $n$ potentially faulty components, $\mf{C} = \{C_1, \ldots, C_n\}$, in the
system. We assume that $\nit{ab}$-atoms, i.e.  propositional variables of the form ${abC_i}$, do not appear in the formulas $\varphi_i$. Accordingly,  no $\nit{ab}$-atom appears negative in the clausal representation of the model. 

This is what makes the model a weak model: nothing is said about how the system behaves under abnormality. This is not the only kind of specifications in CBD. They could also have $\nit{ab}$-atoms in other places of the formulas. However, those we are considering, characterized  as {\em Ignorance under Abnormal Behavior} (IAB),  are common, and have some good properties, among them, that every superset of a minimal diagnosis is a diagnosis. See \cite[sec. 7]{ray92} for a discussion.

 In we also have an observation, $\nit{Obs}$, a set  of non-\nit{ab}-literals, i.e. inputs and outputs, we may consider that the system is not working properly in that:
\begin{equation}
\mc{M} \ \cup \ \nit{Obs} \ \cup \ \{\neg \nit{abC}_i~|~ i = 1, \ldots, n\} \ \mbox{ is inconsistent.} \label{eq:spGen}
\end{equation}

A {\em diagnosis} is a subset {$\Delta$} of {$\{\nit{abC}_1, \nit{abC}_2, \ldots,
\nit{abC}_n\}$}, such that the  assumption that they are true
restores the consistency in (\ref{eq:spGen}), that is,
\begin{equation}
{\mc{M} ~\cup~ Obs ~\cup~ \Delta ~\cup~ \{\neg \nit{abC} ~|~ C \in \mf{C} \mbox{ and } \nit{abC} \notin \Delta\}} \ \mbox{ is consistent}. \label{eq:diag}
\end{equation}

A {\em minimal diagnosis} $\Delta$ is a diagnosis, such that no proper subset of $\Delta$ is also a diagnosis. In general, we want {\em minimal diagnoses}. They are the most informative ones. Sometimes we concentrate on the subclass of {\em minimum diagnoses}, those with minimum cardinality.

Approaching the CBD problem as one about actual causality,
(\ref{eq:spGen}) becomes a natural scenario for performing {\em counterfactual interventions} on the $\neg \nit{ab}$ literals, the only ones declared as endogenous; to detect if they falsify the inconsistency, returning to the expected consistency. On this basis, and according to \cite{HP05}, we can give the following definition.

\begin{definition}\label{def:causes} \em
For a diagnosis setting $\langle\mf{C},\mc{M},\nit{Obs}\rangle$ as in (\ref{eq:spGen}):

\noindent (a) An atom  $\nit{abC}_i$ is a {\em counterfactual cause} \ iff \ $\mc{M} \ \cup \ \nit{Obs} \ \cup \ \{\nit{abC}_i\} \ \cup \ \{\neg \nit{abC}_j~|~ j = 1, \ldots, i-1, i+1, \ldots n\}$ is consistent.

\noindent   (b) An atom $\nit{abC}_i$ is an {\em actual cause}  iff there is $\Gamma \subseteq (\{\nit{abC}~|~C \in \mf{C}\} \smallsetminus \{\nit{abC}_i\})$, such that: \ (b1) \
$\mc{M} \ \cup \ \nit{Obs} \ \cup \ \{\neg \nit{abC}_i\} \ \cup \ \Gamma \ \cup \ \{\neg \nit{abC}~|~ C \in \mf{C}, \nit{abC} \notin (\Gamma \cup \{\nit{abC}_i\})\}$ \mbox{ is inconsistent},
 but \ (b2) $\mc{M} \ \cup \ \nit{Obs} \ \cup \ \{\nit{abC}_i\} \ \cup \ \Gamma \ \cup \ \{\neg \nit{abC}~|~ C \in \mf{C}, \nit{abC} \notin (\Gamma \cup \{\nit{abC}_i\})\}$ is consistent. \ $\Gamma$ is called a {\em contingency set} for $\nit{abC}_i$.

 \noindent (c) The {\em responsibility} of an actual cause $\nit{abC}$ is $\nit{Resp}(\nit{abC}) := 1/(1 + |\Gamma|)$, with  $\Gamma$  a minimum contingency set for $\nit{abC}$. A non-actual cause has responsibility $0$. \boxtheorem
\end{definition}

Notice that the definition can be applied with a model of which we only know the normal input/output behavior. Notice also that every counterfactual cause is also an actual cause with empty contingency set. The following proposition holds.

\begin{proposition} \label{prop:one} \em For a diagnosis setting $\langle \mf{C}, \mc{M}, \nit{Obs}\rangle$: \ (a) A literal $\nit{abC}$ is a counterfactual cause iff $\Delta = \{\nit{abC}\}$ is a diagnosis. \
(b) A literal $\nit{abC}$ is an actual cause with minimal contingency set $\Gamma$ iff $\nit{abC}$ belongs to a minimal diagnosis $\Delta$, and $\Gamma = \Delta \smallsetminus \{\nit{abC}\}$.  \boxtheorem
\end{proposition}

\vspace{-4mm}
\begin{example} \ (example \ref{ex:circ} cont.) \label{ex:cause} \ We had  $
\{a, \neg b, c,\neg d\} \ \cup \ \mc{M} \ \cup \{\neg \nit{abA}, \neg \nit{abO}\}$ as inconsistent.
\ Switching $\neg \nit{abO}$  into $\nit{abO}$, reestablishes consistency:
$\{a, \neg b, c, \neg d\} \ \cup \ \mc{M} \ \cup \{\neg \nit{abA}, \ul{\nit{abO}}\}$ is consistent. Then,
$\nit{abO}$  is a {\em counterfactual cause} for the malfunctioning of the circuit.

  However, when we switch $\nit{abA}$,  \ $\{a, \neg b, c, \neg d\} \ \cup \ \mc{M} \ \cup \{\ul{\nit{abA}},  \neg \nit{abO}\}$ is still inconsistent. Accordingly, $\nit{abA}$ is not a counterfactual cause. It is not an actual cause either: its only potential contingency set $\Gamma = \{\nit{abO}\}$ does not satisfy condition (b1) in Definition \ref{def:causes}(b). \boxtheorem
 \end{example}

\vspace{-3mm}
We restrict Proposition \ref{prop:one} to minimal diagnoses, because for  models of the form (\ref{eq:mod}), every superset of a diagnosis is a diagnosis \cite[sec. 7]{ray92}. However, not every superset of a contingency set is a contingency set for a tuple, due to condition (b1) in Definition \ref{def:causes}.  Furthermore, minimal diagnoses are those of interest.

\vspace{-2mm}
\subsection{Causal Structural Models} Those who are more familiar with causality represented by {\em structural models} \cite{pearl} may be missing them here. Actually, the diagnosis problems can also be cast in those terms. A purely logical model, as in the previous examples, does not distinguish causal directions, or between causes and effects. They can be better represented by a structural model that takes the form of a  {\em causal network}. That is, a directed, acyclic graph whose nodes represent variables; and the directed edges between them represent cause-effect relationships. Those edges may be labeled with deterministic equations or conditional distributions that define  a destination variable in terms of the  variables that point to it \cite{pearl}.

\begin{figure}[h]
 \centerline{\includegraphics[width=7.3cm]{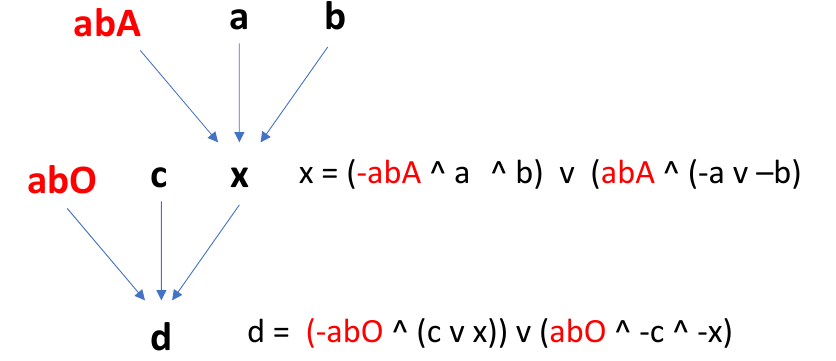}}
 \vspace{-1mm}\caption{Causal Network}\label{fig:CN}\vspace{-6mm}
 \end{figure}

\begin{example} (ex. \ref{ex:circAC} cont.) The causal network in Figure \ref{fig:CN} represents our  possibly faulty circuit. \ In it, $\nit{abA}, \nit{abO}$ are endogenous variables, which can be subject to counterfactual interventions; in this case, making $\nit{abA}$ and $\nit{abO}$ true or false.

 Variables $x$ and $d$ are endogenous, and have {\em structural equations} associated to them, as shown in Fig. \ref{fig:CN}, capturing the circuit's logic. They are used unidirectionally, consistently with the edge directions. In this case, we use by choice \ -contrary to the weak model of failure in Figure (\ref{fig:sat})- \ equations that also specify the behaviour under abnormal conditions.  \boxtheorem
\end{example}

\vspace{-7mm}
 \section{CBD for Explainable Boolean Classification}\label{sec:diagX}

\vspace{-2mm}
 To fix and convey the main ideas, we concentrate on Boolean-circuit classifiers that take Boolean features as inputs, and return a binary label, $1$ or $0$.

\vspace{-2mm}
  \begin{figure}[h]
  \begin{center}
 \includegraphics[width=5cm]{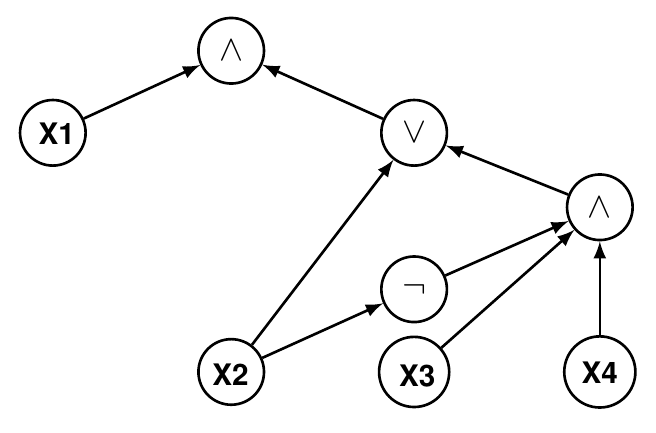}~~~~\includegraphics[width=5.2cm]{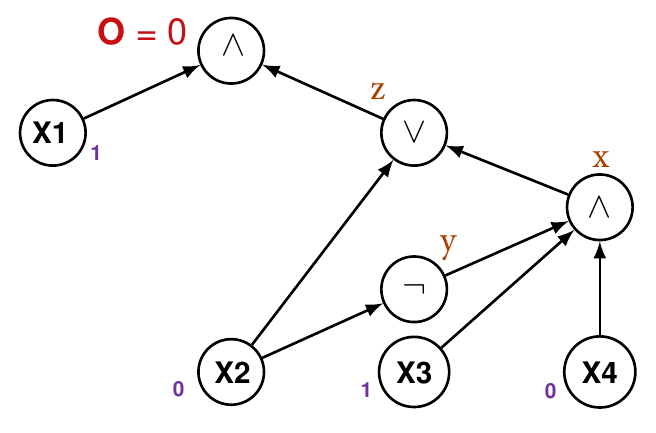}
\vspace{-3mm} \caption{Boolean Classification Circuit}\label{fig:classif}\vspace{-9mm}
  \end{center}
 \end{figure}

 \begin{example} \label{ex:class} Consider the Boolean circuit in Figure \ref{fig:classif}(a).\footnote{A {\em deterministic and decomposable Boolean circuit} (d-DBC) used in \cite{AAAI21}. d-DBCs can encode decision trees, several classes of binary decision diagrams, binary neural networks \cite{jelia}, etc.} The set of propositional input features is $\mc{F} =\{\mbf{x_1}, \mbf{x_2}, \mbf{x_3}, \mbf{x_4}\}$.  The output, $\mbf{O}$,
is at
the top node. The circuit on the right-hand side shows variables for the intermediate gate outputs.
 The BC can be logically specified:  $
\mc{B} = \{
(\neg \mbf{x_2} \longleftrightarrow y), \
(y \wedge \mbf{x_3} \wedge \mbf{x_4} \longleftrightarrow x), \
(\mbf{x_2} \vee x \longleftrightarrow z), \
(\mbf{x_1} \wedge z \longleftrightarrow \mbf{O})$. 
\ As shown in Figure \ref{fig:classif}(b), with  $\mc{B}$ and input entity  $\e$ with:
\begin{equation}
\e(\mbf{x_1}) :=1, \ \e(\mbf{x_2}) :=0, \ \e(\mbf{x_3}) :=1, \ \e(\mbf{x_4}) :=0, \label{eq:signs}
\end{equation}
we obtain
$y = 1, x =0, z=0$, and  the output $\mbf{O}(\e) = 0$.
\boxtheorem \end{example}

\vspace{-3mm}We want to find  input values as actual causes for the observed output, but we want to do this by solving a MBD problem. To approach the problem as CBD, we consider the output as something unexpected under normal circumstances. Since we have no reason to assume that the logical gates of the classifier are working abnormally; we assume they are not faulty. Accordingly,
and departing from Section \ref{sec:cbd}, we allow the  inputs to the classifier to be  faulty in that they do not produce the ``expected output", in this case, $\mbf{\bar{O}} = 1$, the complementary literal of $\mbf{O}$. We denote the original classifier with \ $\mc{B}(\mbf{x_1},\mbf{x_2},\mbf{x_3},$ $\mbf{x_4};\mbf{O})$, indicating that, with inputs $\mbf{x_1},\mbf{x_2},\mbf{x_3},$ $\mbf{x_4}$, it returns -actually, it becomes- the binary value $\mbf{O}$.

\begin{example} (ex. \ref{ex:class} cont.) \label{ex:class2}
We want to explain the output, $\mbf{O}=0$, by identifying the input values that are most relevant for the outcome. As in Section \ref{sec:cbd}, we introduce, for each input features $\mbf{x_i}$, a corresponding propositional variable $\nit{ab}(\mbf{x_i})$, standing for ``$\mbf{x_i}$ takes an abnormal value". With them, we have the base model of failure:
\begin{eqnarray}
\mc{M} \ &:=& \ \mc{B}(\mbf{x_1},\mbf{x_2},\mbf{x_3},\mbf{x_4};\mbf{O}) \cup \{(\neg \nit{ab}(\mbf{x_1}) \longleftrightarrow \mbf{x_1}), \ (\neg \nit{ab}(\mbf{x_2}) \longleftrightarrow \neg \mbf{x_2}), \nonumber \\ && ~~~~~(\neg \nit{ab}(\mbf{x_3}) \longleftrightarrow \mbf{x_3}), \ (\neg \nit{ab}(\mbf{x_4}) \longleftrightarrow \neg \mbf{x_4})\}, \label{eq:notweak}
\end{eqnarray}
where the polarity of the variables on the RHSs of the double implications correspond to those in (\ref{eq:signs}). That set in the second disjunct becomes the ``observed inputs under normal circumstances".

Depending on the truth values of the $\nit{ab}(\mbf{x_i})$ in the double implications, the corresponding values of the associated RHSs, are meant to be the inputs to $\mc{B}(\mbf{x_1},\mbf{x_2},$ $\mbf{x_3},\mbf{x_4};\mbf{o})$. From this model not much can be obtained: We do not have observations for the classifier, everything is conditional. However, the extended model:
\begin{equation}
\mc{T} \ := \ \{\neg \nit{ab}(\mbf{x_1}), \neg \nit{ab}(\mbf{x_2}),\neg \nit{ab}(\mbf{x_3}),\neg \nit{ab}(\mbf{x_4})\} \cup \mc{M} \cup  \{\mbf{\bar{O}}\} \label{eq:inc}
\end{equation}
is inconsistent. In fact, under the assumption of normality, the inputs to the circuit become as in (\ref{eq:signs}); and then, the circuits evaluates to $\mbf{O}=0$, but in $\mc{T}$ we are requesting the output to be its opposite, i.e. $\mbf{\bar{O}} = 1$.

Now, we can again proceed as in Section \ref{sec:cbd},
obtaining from $\mc{T}$ diagnoses that contain abnormality propositional atoms.
Accordingly, $\Delta \subseteq \{ \nit{ab}(\mbf{x_i})~|~\mbf{x_i} \in \mc{F}\}$ is a diagnosis if, changing in $\mc{T}$ the $\neg \nit{ab}(\mbf{x_i})$ into $\nit{ab}(\mbf{x_i})$ when $\nit{ab}(\mbf{x_i}) \in \Delta$,
restores consistency.

Notice that due to the formulas $(\neg \nit{ab}(\mbf{x}_1) \longleftrightarrow \mbf{x_1})$, etc., declaring $\nit{ab}(\mbf{x}_i)$ to be true also changes the value of $\mbf{x_i}$ to its inverse, which is what we do in actual causality.
\ With $\Delta = \{\nit{ab}(\mbf{x_2})\}$, that is, changing in $\mc{T}$, $\neg \nit{ab}(\mbf{x_2})$ into $\nit{ab}(\mbf{x_2})$, we obtain the new input value $\mbf{x_2} = 1$, and we obtain $\mc{B}(1,\underline{1},1,0;\underline{\mbf{1}})$, whose output does not collide with the intended one at the very right of (\ref{eq:inc}).  $\Delta$ is a diagnosis.
\boxtheorem \end{example}

\vspace{-3mm}Notice that, due to  the double implications in (\ref{eq:notweak}), model $\mc{M}$  is not a {\em weak} model of failure anymore: $\mc{M}$ also tells us under what conditions there is abnormality, and not only what happens when there is no abnormality. Using only left-to-right arrows in (\ref{eq:notweak}) would not produce the intended changes of input values. If we want a weak model, we could drop the right-to-left arrows, modifying the definition of diagnosis, as follows: $\Delta \subseteq \{ \nit{ab}(\mbf{x_i})~|~\mbf{x_i} \in \mc{F}\}$ is a diagnosis if, changing in $\mc{T}$ the $\neg \nit{ab}(\mbf{x_i})$  into $\nit{ab}(\mbf{x_i})$ \ {\bf and} \ the associated inputs $\mbf{x_i}$ into their inverses,
restores consistency.

 Notice that the method just presented does not need the internals of the classifier, and could be applied with a black-box binary classifier $\mc{C}(\mbf{x_1},\mbf{x_2},\mbf{x_3},\mbf{x_4};\mbf{O})$.\\

\section{Exploiting the Connections}\label{sec:connections}

In this section, we show and example of how known techniques and results for CBD can be used to investigate applications of actual causality and resposibility. This, in the context of Explainable Data Management, where, more specifically, we want to explain how a Boolean query $\mc{Q}$, successfully answered by a database $D$, becomes true.

In this case, we want explanations in terms of {\em which} DB tuples contribute to the positive answer, and by {\em how much}. The latter is answered by means of an attribution score, which, in our case, turns out to be causal responsibility (for more on the subject and other attribution scores in XDM, see \cite{adbis23}).
\ For illustration purposes, we consider Boolean conjunctive queries, i.e. of the form $\mc{Q}\!: \bar{\exists} (P_1(\bar{x}_1) \wedge \cdots \wedge P_n(\bar{x}_n))$, which is fully existentially quantified.

\begin{example}\label{ex:mbdaex6}
\ Consider the database $D = \{ R(c, b), R(a, d), R(b, a), R(e, f), S(a),$ $ S(b),  S(c), S(d)\}$, whose elements are called tuples.  Every ground atom written in the language of $D$'s schema, can be seen as a propositional variable, which is true when $\tau \in D$, and false otherwise.

Now, the query $\mc{Q}\!: \exists x \exists y ( S(x) \land R(x, y) \land S(y))$ becomes
true in $D$; and  the join in it can be satisfied with different combinations of tuples in $D$. For example, by the tuples $S(c), R(c,b), S(b)$. We want to identify the tuples that are actual causes for $\mc{Q}$ to be true. For example, if
$S(b)$ is deleted from $D$, as an intervention, this particular instantiation of the join becomes false.

In more general terms, $\mc{Q}$ is true in $D$, and we want to {\em invalidate} $\mc{Q}$ by intervening tuples; here, by deleting tuples (adding tuples to $D$ will not invalidate a conjunctive query). By doing so, we can identify tuples as actual causes. However, instead of  tuple interventions in relation to the query at hand, we can reduce the problem to a CBD problem.

Since we want to {\em invalidate the query}, the query is transformed into its negation, which becomes a    {\em denial integrity constraint} on $D$, namely, \ $\kappa\!: \ \neg \exists x \exists y (S(x) \wedge R(x, y)  \wedge \ S(y))$, which prohibits the satisfaction of the query join. \
 It holds that $\mc{Q}$ is satisfied by $D$ iff $\kappa$ is violated by $D$, which should be considered the faulty behavior under normal conditions. \ Accordingly, we specify a  weak model of failure:
\begin{eqnarray}
D \ \cup \ \{\forall x \forall y(\neg\nit{AbS}(x) &\wedge& \neg\nit{AbR}(x,y) \wedge \neg \nit{AbS}(y) \ \longrightarrow \label{eq:dbMod}\\
&&~~~~~~~~~~((S(x)  \wedge R(x, y)  \wedge S(y)) \rightarrow \nit{false})\}. \nonumber
\end{eqnarray}

The model includes an abnormality predicate for each relational predicate, and an always false propositional atom, $\nit{false}$. Formula \ref{eq:dbMod} tells us that, when the tuples are not abnormal, they do not participate in the violation of the IC $\kappa$. That is,
under normality assumptions, the database does not make the query true.

Here, the observation is $\mc{Q}$ itself, which, being true in $D$, combined with (\ref{eq:dbMod}) produces an inconsistent theory, when the  $\nit{AbS}$- \ and $\nit{AbR}$-atoms in (\ref{eq:dbMod}) are all false. According to CBD, for the combinations of $S$- and $R$-atoms that make the join true, some of those abnormality atoms  have to be true.  In particular,
at least one of  $\nit{Ab}_S(c)$, $\nit{Ab}_R(c,b)$, $ \nit{Ab}_R(b)$ has to be true.
\ The tuples whose associated abnormality atoms become true are the actual causes for the query.

From the minimal diagnoses that contain a tuple, we can compute minimal contingency sets for it, and eventually, its responsibility. \boxtheorem
\end{example}

\vspace{-3mm}
In \cite{tocs}, this kind of  reduction from actual causality for QA in DBs to CBD turned out to be useful to obtain algorithmic and complexity results for responsibility  in DBs, and for CBD as well. This was achieved via algorithms based on computing diagnoses as hitting-sets of {\em conflicts} in CBD \cite{reiterDiag};\footnote{A {\em conflict} in CBD is a set of negative \nit{ab}-literals whose conjunction is inconsistent with the CBD model plus the observation.} and also via results for minimum-size {\em DB repairs} for the inconsistency associated to the denial constraints \cite{CQA,icdt07}: minimum-size due to the need for minimum-size contingency sets that underlie responsibility.

\vspace{-1mm}
\section{Conclusions}\label{sec:conclusions}

\vspace{-1mm}
We have barely started to scratch the surface of the connections between consist-ency-based diagnosis and actual causality, and their applications to XAI and XDM. The connections are interesting {\em per se}, and could be extended in different directions. Going beyond the propositional setting would be interesting, and useful. Related to this, in \cite{deem} (see also \cite{adbis23}), the responsibility attribution score was extended to the non-binary case. The extension is not trivial, and deserves more investigation. Similar challenges should appear in the connection to CBD.

Several computational techniques and results have been introduced and established in actual causality {\em cum} responsibility, on one side, and CBD on the other; following independent paths. For example, it would be interesting to investigate the meaning and applicability of the notion of {\em kernel diagnosis} \cite{ray92} when applied to actual causality.  It would also be interesting to investigate its connections to the {\em core} of database repairs \cite{nina}, and closer connections  with sufficient (or abductive) explanations \cite[sec. 6]{ray92}.

By applying actual causality in a CBD setting, we have been able to use the quantitative responsibility score to identify the most relevant (elements) of diagnoses.  It would be interesting to develop algorithms for computing early, or only, highly relevant diagnoses or elements thereof. Such an idea was already proposed in \cite[sec. 8]{ray92}, but without relation to responsibility. Instead they propose using additional domain or probabilistic knowledge. This could also be interesting in a setting where responsibility is applied.\footnote{For responsibility under database integrity constraints, which goes along these lines, see \cite{flairsExt,bda22}.}

\vspace{3mm}
\noindent {\bf Acknowledgements:}   L. Bertossi has been financially supported by the IMFD, Chile; and NSERC-DG 2023-04650, Canada. Useful comments from the reviewers are much appreciated.

\bibliographystyle{plain}

\end{document}